\documentclass[journal]{IEEEtran}

\usepackage{xcolor,soul,framed} 
\colorlet{shadecolor}{yellow}
\usepackage[pdftex]{graphicx}
\graphicspath{{../pdf/}{../jpeg/}}
\DeclareGraphicsExtensions{.pdf,.jpeg,.png}
\usepackage{subcaption}
\usepackage{adjustbox}
\usepackage{booktabs, siunitx}

\usepackage{array}
\usepackage{mdwmath}
\usepackage{mdwtab}
\usepackage{eqparbox}
\usepackage{url}
\usepackage{cite}
\usepackage{multirow}
\usepackage{commath}
\usepackage{longtable}
\usepackage{relsize}
\usepackage{amssymb}
\usepackage{amsmath, bm}
\usepackage{bbm}
\usepackage{amsfonts} \usepackage[switch]{lineno}
\usepackage{parskip}
\usepackage{rotating}

\begin{document}
\bstctlcite{IEEEexample:BSTcontrol}
    \title{TopoCL: Topological Contrastive Learning for Time Series}
\author{
    \IEEEauthorblockN{Namwoo Kim\IEEEauthorrefmark{1}, Hyungryul Baik\IEEEauthorrefmark{2}, Yoonjin Yoon\IEEEauthorrefmark{1}}
    \\\IEEEauthorblockA{\IEEEauthorrefmark{1} Department of Civil and Environmental Engineering, Korea Advanced Institute of Science and Technology
    \\\{ih736x, yoonjin\}@kaist.ac.kr}
    \\\IEEEauthorblockA{\IEEEauthorrefmark{2}Department of Mathematical Science, Korea Advanced Institute of Science and Technology
    \\\{hrbaik\}@kaist.ac.kr}
}


\markboth{IEEE Transactions on Neural Networks and Learning Systems
}{}


\maketitle

\begin{abstract} 
Universal time series representation learning is challenging but valuable in real-world applications such as classification, anomaly detection, and forecasting. Recently, contrastive learning (CL) has been actively explored to tackle time series representation. However, a key challenge is that the data augmentation process in CL can distort seasonal patterns or temporal dependencies, inevitably leading to a loss of semantic information. To address this challenge, we propose Topological Contrastive Learning for time series (TopoCL). TopoCL mitigates such information loss by incorporating persistent homology, which captures the topological characteristics of data that remain invariant under transformations. In this paper, we treat the temporal and topological properties of time series data as distinct modalities. Specifically, we compute persistent homology to construct topological features of time series data, representing them in persistence diagrams. We then design a neural network to encode these persistent diagrams. Our approach jointly optimizes CL within the time modality and time-topology correspondence, promoting a comprehensive understanding of both temporal semantics and topological properties of time series. We conduct extensive experiments on four downstream tasks—classification, anomaly detection, forecasting, and transfer learning. The results demonstrate that TopoCL achieves state-of-the-art performance.
\end{abstract}


%


\section{Introduction}

\IEEEPARstart{T}ime series analysis plays a crucial role in numerous real-world applications such as weather, economics, and transportation. However, analyzing time series data is challenging due to its incompleteness, vulnerability to noise, and complexity. Additionally, employing deep learning models for time series analysis introduces further obstacles, as it necessitates well-labeled data. Labeling time series data is time-consuming and requires a high level of expertise, creating significant bottlenecks in the analysis process. 
In recent years, contrastive learning (CL) has become increasingly popular in computer vision (CV)~\cite{chen2020simple,he2020momentum} and natural language processing (NLP) ~\cite{gao2021simcse}, mainly due to its ability to train models without explicit labels. Accordingly, ongoing research is focused on applying CL to time series data for various downstream applications, including anomaly detection~\cite{yue2022ts2vec,liu2024timesurl}, forecasting~\cite{woo2022cost}, and classification ~\cite{TSTCC}.

In general, CL leverages data augmentation to create positive and negative pairs for training, thereby creating a contrast between similar and dissimilar samples by exploring different transformations and variations of the data. Several augmentation techniques for time series have been proposed to learn robust and discriminative representations. For example, $permutation$ reorders segments within a time series to produce a new sequence ~\cite{TSTCC, DBLP:conf/aaai/MengQLCX023}. However, using $permutation$ on time series data might introduce inappropriate inductive bias, as it can distort the autocorrelated nature of time series data. Techniques borrowed from CV, such as $flipping$, which flips the signs of the original time series, may misrepresent the trend or seasonal patterns. Moreover, these distortions raise concerns about whether the augmented time series are indeed positive samples of the original series. Consequently, applying data augmentation may inevitably risk information loss, making it difficult to capture semantically meaningful representations. Therefore, it is necessary to find ways that compensate such potential information loss.

Recently, Topological Data Analysis (TDA) has emerged as a subfield of algebraic topology focused on capturing the global \textit{shape} of data, where \textit{shape} broadly refers to data properties that remain invariant under transformations ~\cite{carlsson2009topology}. Such shape patterns, or topological features, include connected components, loops, and voids. Topological features are obtained by computing persistent homology and are represented as a multiset in persistence diagrams (PD) ~\cite{cohen2005stability}. These diagrams encapsulate persistent homology across different scales of data. The overall structure and distribution of sets in the diagram provide insights into the topological invariants of the data. By incorporating such topological information, we can capture the essential structural properties of the time series that remain invariant under various transformations. Therefore, integrating topological information can help compensate for potential information loss due to data augmentation, reduce inductive bias, and lead to more robust representations.


Fueled by the advancements in both CL and TDA, we harness the strengths of the two methods and introduce Topological Contrastive Learning for time series (TopoCL), a simple yet effective cross-modal CL approach for universal time series representations. Our method integrates and leverages information from the time and topology modalities. To start, we construct topological characteristics from time series data. We apply delay embedding to time series data, compute persistent homology, and design a simple topological module to effectively encode topological information. Our approach then focuses on a joint objective: ensuring that the augmented versions of the same time series are closely embedded together in the feature space while preserving the correspondence between time and topology. The joint learning objective promotes the comprehensive understanding of temporal semantics and topological properties of time series and improves robustness to data augmentations. We perform extensive experiments on time series forecasting, anomaly detection, classification, and transfer learning tasks. The experimental results demonstrate that the learned representations of TopoCL are effective. The contributions of this work are as follows.

\begin{itemize}
    \item We demonstrate that incorporating topological properties is effective for capturing robust and discriminative representations of time series data.
    
    \item We design a novel framework for learning representations of time series data that takes into account their topology. To the best of our knowledge, our approach is the first to combine persistent homology with contrastive learning in the context of time series representation learning.

    \item  
    We conduct extensive experiments across four downstream tasks: time series classification, anomaly detection, forecasting, and transfer learning on diverse datasets. The experimental results demonstrate the effectiveness of the proposed model.
\end{itemize}

The remainder of this paper is organized as follows. Section 2 presents a literature review on TDA and time-series contrastive learning. Section 3 provides preliminaries. The proposed framework is described in Section 4. Section 5 provides the experimental results of the proposed model. Section 6 presents additional analysis. Finally, Section 7 presents the conclusions and future work.

\section{Related Work}
\subsection{Topological Data Analysis}
TDA is an emerging field that utilizes abstract algebra to uncover intrinsic shape of the data. A fundamental concept in TDA is PD, which helps to understand the topological structure of data. By computing persistent homology across scales, such topological structure is represented as a multiset of points. Directly using PD in machine learning and deep learning models is challenging, since these models typically work with Hilbert space. To overcome this, several research efforts have been made to convert PD into vector format. For instance, persistence landscapes transform the multiset of points in a PD into a collection of piecewise-linear functions~\cite{bubenik2015statistical}.~\cite{adams2017persistence} proposed persistence images, which converts PD into a fixed-size representation. Additionally, various kernel functions have been proposed to handle PD, such as geodesic topological kernel~\cite{padellini2021supervised} and persistence landscape-based kernels~\cite{zhu2016stochastic, bubenik2020persistence}.

Recently, the integration of TDA into neural networks has been explored, leading to the development of various topological layers for machine learning applications. The first approach to input a PD into a neural network architecture was presented by Hofer \textit{et al.} \cite{hofer2017deep}. Carrière \textit{et al.} \cite{carriere2020perslay} proposed PersLay, which incorporates PD for graph classification. Moor \textit{et al.} \cite{moor2020topological} proposed topological autoencoder preserving topological structures of the input space in latent representations. Kim \textit{et al.} \cite{kim2020pllay} proposed PLlay, a neural network layer for learning embeddings from persistence landscapes. These approaches demonstrated that incorporating persistent homology with deep learning has the potential to offer a more comprehensive understanding of data. 

\subsection{Time Series Contrastive Learning} 
Similar to the advancements of CL in the fields of CV and NLP, numerous studies incorporate CL into time series analysis. TS-TCC ~\cite{TSTCC} employed both weak and strong augmentations to time series data and utilized contextual contrasting to learn transformation-invariant representations. Mixing-up ~\cite{wickstrom2022mixing} generated an augmented sample by mixing two data samples and proposed a pretext task method for predicting the mixing proportion. Inspired by great success in masked modeling in NLP and CV, several works adopted this approach for time series. TS2vec~\cite{yue2022ts2vec} used masking and cropping for data augmentation, and proposed a hierarchical contrastive loss to learn scale-invariant representations. SimMTM~\cite{dong2024simmtm} generated multiple masked series and facilitated reconstruction by assembling complementary temporal variations from multiple masked series. InfoTS~\cite{luo2023time} exploited a meta-learning framework to automatically select the augmentation method.

Recently, several studies have focused on the use of the frequency domain of time series data. CoST ~\cite{woo2022cost} proposed disentangled seasonal-trend representation learning framework, incorporating contrastive loss in both the time and frequency domains. TF-C~\cite{zhang2022self} proposed a novel contrastive loss to maintain consistency between frequency and time domain representations. TimesURL~\cite{liu2024timesurl} proposed a novel frequency-temporal-based augmentation method to maintain temporal dependencies of time series. However, it still remains unclear whether the proposed methodologies effectively mitigate the inevitable information loss resulting from data augmentation. To address this issue, we propose using persistent homology to capture the essential structural properties of time series data, thereby mitigating such information loss.

\section{Preliminary}
In this section, we give the formal definitions of the major terminologies of TDA used in this study. 

\subsection{Delay Embedding} 
Takens's delay embedding \cite{takens2006detecting} is a commonly used technique for converting time series data into point cloud representations.  Given an $i-$th time series $x_{i}=\{x_{i,1}, x_{i,2}, \dots, x_{i,T}\}$, one can extract a sequence of vectors of the form $[x_{i,t}, x_{i,t+\gamma}, \dots ,x_{i,t+(m-1)\gamma}$ $]\in$ $\mathbb{R}^m$, where $m$ is the embedding size and $\gamma$ is the time delay.
\subsection{Simplicial Complex}
In algebraic topology, a simplex is a geometric object that generalizes the notion of a triangle or a tetrahedron to higher dimensions. Specifically, a $k-$simplex is defined as a convex hull spanned by $k+1$ points $v \in \mathbb{R}^m$ that are affinely independent. Then, a simplicial complex $\mathcal{K}$ is a finite collection of simplices such that:
\begin{itemize}
    \item If $\sigma$ $\in$ $\mathcal{K}$, every subset of $\sigma$ (i.e., every face of $\sigma$) is an also element of $\mathcal{K}$
    \item If two simplices $\sigma_1$, $\sigma_2$ $\in$ $\mathcal{K}$ intersects, their intersection is a face of both $\sigma_1$ and $\sigma_2$.
\end{itemize}

\subsection{Vietoris-Rips Complex}
To construct a simplicial complex, we used the Vietoris-Rips complex \cite{vietoris1927hoheren}. Let $(X, d)$ be a finite metric space equipped with a distance function $d$. The Vietoris-Rips complex $VR(X, \epsilon)$ is the abstract simplicial complex whose simplices are subsets of $X$ with diameter (i.e., the maximum pairwise distance between points) at most $\epsilon \ge 0$. That is, a subset $S$ of $X$ is a simplex in $VR(X, \epsilon)$ if and only if the distance $d(x,y) \le \epsilon$ for any two points $x, y\in S$ . The $0$-simplices of $VR(X, \epsilon)$  are the points of X, and the $k$-simplices are the subsets of $X$ of cardinality $k+1$ that can be realized as the union of $k+1$ closed balls of radius $\epsilon$ with non-empty pairwise intersections. 

\subsection{Homology}

\subsubsection{Simplicial Homology}
Given a simplicial complex $\mathcal{K}$, a weighted sum of $n$-simplices with coefficients from $\mathbb{Z}/\mathbb{Z}_2$-defines an $n$-chain on $\mathcal{K}$. Then the space of $n$-chains on $\mathcal{K}$, represented by $C_n(\mathcal{K})$, is defined as the vector space spanned by $n$-simplicies of $\mathcal{K}$ over $\mathbb{Z}/\mathbb{Z}_2$. The modules $C_0, C_1, ...$ are connected by boundary operators $\partial_n:C_n(\mathcal{K}) \rightarrow C_{n-1}(\mathcal{K})$. The boundary operator for a simplex $\sigma=[v_0, ..., v_n]$ is defined as: 

\begin{align}
    \partial _n (\sigma) = \sum_{i=0}^{n} [v_0, ..., \hat{v_i} ,..., v_n]
\end{align}where $[v_0, ..., \hat{v_i} ,..., v_n]$ denotes the $(n-1)$ simplex spanned by all the vertices except $v_i$. It is easy to see that $im(\partial _{n+1})$ is contained in $ker(\partial _n)$ for each $n$. The $n$-th homology group of $\mathcal{K}$ is defined as:

\begin{align}
    H_n(\mathcal{K}) = ker(\partial _n)/im(\partial _{n+1}).
\end{align}%
The $n$-th Betti number of $\mathcal{K}$, denoted as $\beta_n(\mathcal{K})$, is the dimension of the $n$-th homology group of $\mathcal{K}$, which can be computed as the difference between the kernel dimension of $\partial_n$ and the image dimension of $\partial_{n+1}$. In simpler terms, the $n$-th Betti number indicates the number of $n$-dimensional voids that exist in the simplicial complex. For example, $\beta _0$ indicates the number of connected components, while $\beta_1$ represents the number of loops.

\subsubsection{Persistent Homology}
Persistent homology is a valuable technique in TDA for understanding the shape of complex data sets. For a given simplicial complex $K$, let $(K)_{i=0}^m$ be a nested subcomplexes of $K$ such that $\varnothing=K_0\subseteq K_1 \subseteq ... \subseteq K_m=K$, called a \textit{filtration}. The inclusion maps $K_i \hookrightarrow K_j$ defined by $f(x)=x$ induce $f_{i,j}: H_{p}(K_{i})\to H_{p}(K_{j})$ for all $i,j$ with $1\leq i\leq j\leq m$, for all $p$. Then, the sequence of homology groups is represented as follows
\begin{align}
    0 = H_p(K_0) \rightarrow H_p(K_1) \rightarrow  \ldots \rightarrow H_p(K_m)=H_p(K).
\end{align}

For given $i \leq j$ and $p$, we can consider the $p$-th boundary maps $\partial_p^i : C_p(K_i) \to C_{p-1}(K_i)$ and $\partial_p^j : C_p(K_j) \to C_{p-1}(K_j)$. Since $K_i \subset K_j$, one can see $ C_p(K_i)$ and $C_{p-1}(K_i)$ as subspaces of $C_p(K_j)$ and $C_{p-1}(K_j)$, respectively. Then both $im (\partial_{p+1}^j) \subset C_{p}(K_j)$ and $ker (\partial_p^i) \subset C_p(K_i)$ can be considered as subspaces of $C_p(K_j)$. Then $p-$th persistent homology group is defined as:

\begin{equation}
    H_p^{i,j} = ker (\partial_p^{i}) / (im (\partial_{p+1}^j) \cap ker (\partial_p^i))
\end{equation}%

Note that an element of $ker (\partial_p^{i})$ represents both an element of $H_p(K_i)$ and an element of $H_p(K_j)$. If it vanishes in $H_p^{i,j}$, then it must be in $im (\partial_{p+1}^j)$, hence it vanishes in $H_p(K_j)$ as well. In summary, the homology group $H_p^{i,j}$ consists of the $p$-th homology classes of $K_i$ that are still present at $K_j$, i.e, those not in the kernel of the map $f_{i,j}$. 

The $p$-th Betti number, which refers to the rank of the homology group, is defined as: $\beta _p^{i,j} := \operatorname{rank} H_{p}^{i,j}$.  By analyzing the sequence of Betti numbers, persistent homology enables the systematic measurement of topological features (e.g., holes and voids) in dataset across multiple scales. Such topological information can be summarized as PD.

\subsection{Persistence Diagram}
In TDA, PD is a useful tool for summarizing topological information from data across different scales. A PD captures topological features such as connected components and loops, showing when these features appear and disappear as the scale parameter changes.
The $p$-th PD $\operatorname{Dgm}_p$ is defined as a multi-set collection of points $(i,j)$ with $\mu_p^{i,j}=1$, where $\mu_p^{i,j}$ is calculated as:

\begin{equation}
    \mu_p^{i,j} = (\beta _p^{i,j-1}-\beta _p^{i,j}) - (\beta _p^{i-1,j-1}-\beta _p^{i-1,j})
\end{equation}

\begin{figure}[]
    \centering
        \begin{subfigure}[b]{0.45\textwidth}
        \centering
        \includegraphics[width=0.85\linewidth]{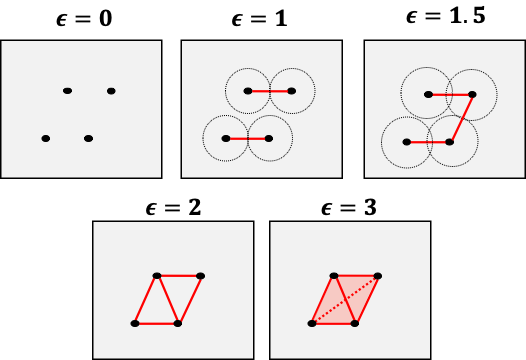}
        \caption{Vietoris-Rips filtration}
        \vspace{0.3cm}
        \end{subfigure}
        \begin{subfigure}[b]{0.45\textwidth}
        \centering
        \includegraphics[width=0.6\linewidth]{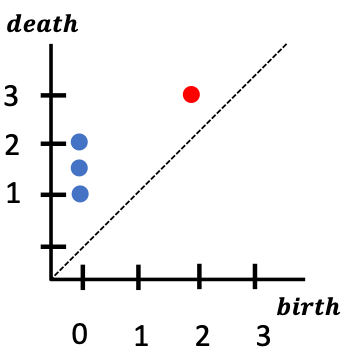}
        \caption{Corresponding persistence diagram}
        \end{subfigure}
    \caption{Illustration of persistence diagram}
\label{fig:vrpd}
\end{figure}

Each point $(a, b)\in \operatorname{Dgm}_p$ corresponds to a homological feature that is born at scale $a$ and dies at scale $b$. This homological feature has the persistence value of $b-a$. Figure ~\ref{fig:vrpd} illustrates an example of PD based on the Vietoris-Rips filtration. The diagram effectively summarizes evolving homological information with respect to $\epsilon$. For instance, as $\epsilon$ increases 0 to 1, the number of connected components decreases 4 to 2, which is reflected as a point (0,1) on the persistence diagram.

\section{Proposed TopoCL Framework}
In this section, we present \textit{TopoCL} framework. 

\subsection{Problem Formulation}
Our goal is to train a neural network $f_\theta$ to transform each time series instance $x_i \in \mathbb{R}^{T\times C}$ in the time series set $X_{time}=\{x_1, x_2, ..., x_N \}$ into its representation $r_i$. Here, $N$ denotes the number of samples, $T$ denotes the number of timestamps, $C$ represents the number of variables. The representation $r_i$ is expressed as $\{r_{i,1}, r_{i,2}, ..., r_{i,T}\}$, where $r_{i,t}\in \mathbb{R}^F$ is the embedding vector at time $t$, with $F$ being the dimension of embedding vector.

\subsection{Overview}
\begin{figure*}[h]
\begin{subfigure}[t]{0.73\linewidth}
\includegraphics[width=0.85\linewidth]{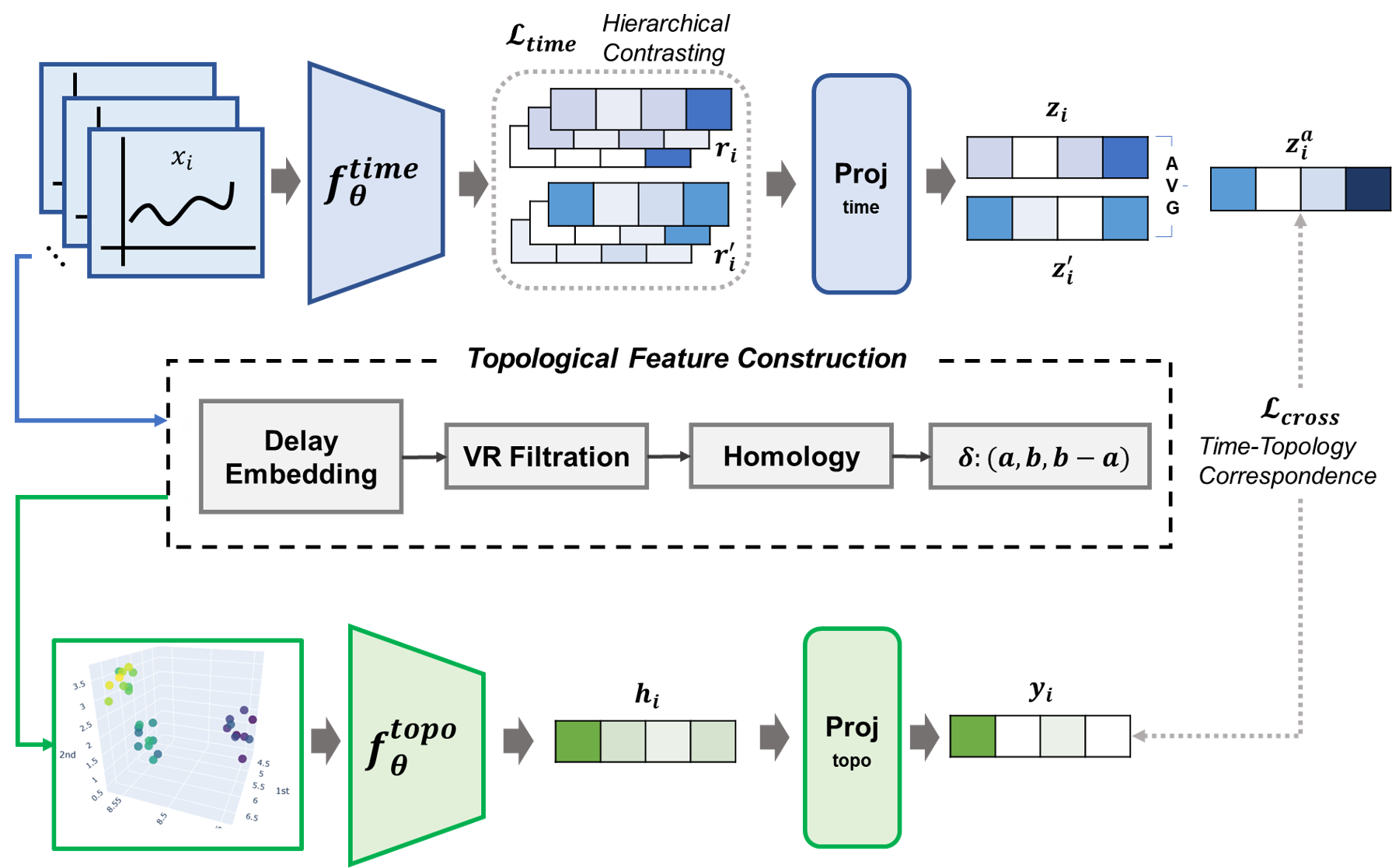}
\caption{}
\end{subfigure}
\hspace{0.05\linewidth}
\centering
\begin{subfigure}[t]{0.19\linewidth}
\includegraphics[width=0.85\linewidth]{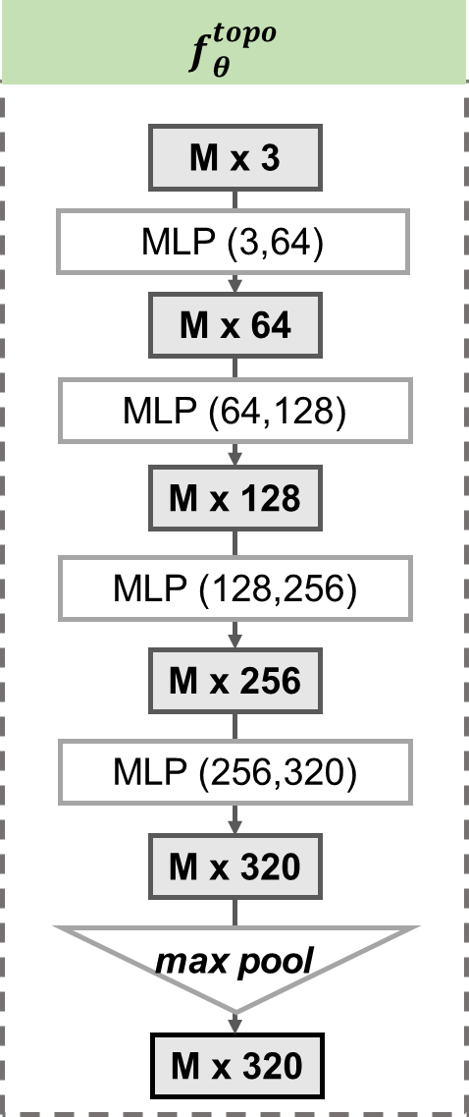}
        \caption{}

\end{subfigure}
\caption{(a) Overall framework of TopoCL. TopoCL consists of two distinct modules: a temporal module and a topological module. The temporal module learns semantically meaningful temporal characteristics using a hierarchical contrastive loss, while the topological module facilitates time-topology correspondence through cross-modal contrastive loss. To achieve this, topological feature construction is performed in three steps: time delay embedding, persistence diagram calculation via Vietoris-Rips filtration, and point cloud transformation. Subsequently, a contrastive loss between the embeddings from the time and topology modalities is applied to ensure time-topology correspondence. TopoCL jointly optimizes the learning process through both intra-modal and cross-modal correspondences. After pre-training, the embedding vectors obtained from $f^{time}_\theta$ are used for downstream applications. (b) Topological feature extractor $f^{topo}_\theta$. Topological feature extractor comprises a series of multilayer perceptrons with ReLU activations and max-pooling layer.}
\label{fig:framework}
\end{figure*}

The overall framework of TopoCL is shown in Figure \ref{fig:framework}. Given $X_{time}$, we construct the topological modality $X_{topo}$ by computing its persistence diagram. Then, $X_{time}$ is fed into the \textit{temporal module} to capture temporal dependencies. Concurrently, $X_{topo}$ is processed by the \textit{topological module} to learn the underlying topology of the data. Temporal features are learned using a time modality contrastive loss $\mathcal{L}_{time}$. Additionally, an auxiliary contrastive objective $\mathcal{L}_{cross}$ is applied to align temporal features and topological features. The output of the temporal encoder $f^{time}_{\theta}$ is used for downstream applications.

\subsection{Persistence Diagram Construction}
To extract topological features from a time series, we first convert $x \in X_{time}$ into a point cloud format, enabling the construction of the Vietoris-Rips filtration. This involves applying delay embedding to the input sample $x$. Next, we compute persistence diagrams for $H_0$ and $H_1$ homology groups, summarizing the topological features of the data. For each channel $c$ in time series data, we obtain channel-specific $0$- and $1$-dimensional persistence diagrams $\mathrm{Dgm}^{c}(x)$. A composite diagram $\mathrm{Dgm}(x)$ is then created as $\mathrm{Dgm}(x) = \cup_c \mathrm{Dgm^c}(x)$. 
Next, points in $\mathrm{Dgm}(x)$ are mapped using $\delta:(a,b) \rightarrow (a, b, b -a)$, resulting in point cloud data $x^p \in \mathbb{R}^{M\times 3}$, where $M$ is the maximum number of topological features. The set of topological feature is denoted as $X_{topo}=\{x^p_1, x^p_2, ..., x^p_N\}$. Further details of topological feature construction is provided in appendix.

\subsection{Topological Feature Extraction}
We design a topological feature extractor $f^{topo}_\theta$, which directly processes unordered point sets $x^p$ as inputs. Given that a point cloud is an unordered collection of points, a neural network capable of processing these $M$ point sets needs to be invariant to $M!$ possible permutations of the input set's order during data feeding.

Inspired by the works on PointNet~\cite{qi2017pointnet} and Deep Sets~\cite{zaheer2017deep}, we apply a simple symmetric function to combine information from all the points in $x^p$. Then, the model is unaffected by the order of input points, preserving invariance to input permutations. As illustrated in Figure \ref{fig:framework}
, $x^p$ passes through the series of shared multilayer perceptrons with ReLU activation. Subsequently, a max-pooling layer aggregates information from all points to produce a topological representation vector $h\in\mathbb{R}^H$, where $H$ is the dimension of representation vector.

\subsection{Temporal Feature Extraction}
Capturing the temporal dependency is essential for time series representation learning. Given an input instance $x_i$, we construct augmented versions of it. Then, the time series feature extractor $f^{time}_\theta$ maps them to a feature embedding space, resulting in embeddings $r_{i}$ and $r_{i}^{\prime}$. 

To learn discriminative and robust temporal representations, we use temporal and instance-wise contrastive losses, $\ell_{\text{temp}}^{(i, t)}$ and $\ell_{\text{inst}}^{(i, t)}$ \cite{yue2022ts2vec}. For a given timestamp $t$, these two loss functions can be formulated as:
\begin{equation}
    \ell_{\text{temp}}^{(i, t)}=-\log \frac{\exp (r_{i, t} \cdot r_{i, t}^{\prime})}
    {\sum \limits_{t'}({\exp{(r_{i, t}\cdot r'_{i, t})}} + \mathbbm{1}_{[t\neq t']} {\exp{\left(r_{i, t}\cdot r_{i, t}\right)}})}
\label{temploss}
\end{equation}

\begin{equation}
    \ell_{\text{inst}}^{(i, t)}=-\log \frac{\exp (r_{i, t} \cdot r_{i, t}^{\prime})}
    {\sum \limits_{j}({\exp{(r_{i, t}\cdot r'_{j, t})}} + \mathbbm{1}_{[i\neq j]} {\exp{\left(r_{i, t}\cdot r_{j, t}\right)}})}
\label{instloss}
\end{equation}
Time modality contrastive loss function $L_{time}$ for a minibatch $B$ is defined as 
\begin{equation}
\mathcal{L}_{\text{time}}=\frac{1}{BT} \sum_{i} \sum_{t}\left(\ell_{\text {temp }}^{(i, t)}+\ell_{\text {inst }}^{(i, t)}\right)
\label{dualloss}
\end{equation}
Next, we apply the hierarchical contrasting method proposed by \cite{yue2022ts2vec}. This involves performing max pooling on $r_{i, t}$ and $r_{i, t}^{\prime}$ along the time axis, and then recursively compute Equation \ref{dualloss}.
\subsection{Time-Topology alignment}
In addition to $\mathcal{L}_{time}$, which learns instance-specific characteristics and temporal variation, we propose a contrastive objective for alining time and topology modalities. To this end, we first map the embedding vectors $r_{i,t}, r_{i,t}^{\prime}\in \mathbb{R}^F$ from temporal encoder $f^{time}_\theta$, and topological features $h_i \in \mathbb{R}^H$ to the multi-modal latent space. Then, we apply projection headers $proj_{time}$ and $proj_{topo}$as follows:
\begin{equation}
    y_i = proj_{p}(h_i)
\end{equation}
\begin{equation}
    z_i = proj_{t}(r_{i}^*) 
\end{equation}
\begin{equation}
    z'_i = proj_{t}(r_{i}^{*\prime})) 
\end{equation}
where, $r_{i}^*$ is an instance level representation of $r_{i,t}$ which is obtained by applying max-pooling across all timestamps. The average of $z_i$ and $z'_i$ is computed as:
\begin{equation}
    z^{a}_i = \frac{1}{2} (z_{i} +z'_{i})
\end{equation}
Next, we maximize the similarity between $z^{a}_i$ and $y_i$, which are mapped into the multi-modal latent space. The similarity between $z_i^a$ and $y_i$ is defined as $ s(z_i^a, y_i) = \frac{z_i^a \cdot y_i}{\norm{z_i^a} \norm{y_i}}/\tau$, where $\tau$ refers to the temperature parameter. For a mini-batch $B$, ($z^{a}_i$, $y_i$) is regarded as a positive pair, where as the remaining $2B-2$ feature vectors constitute negative pairs. Our cross-modal contrastive loss between $z^{a}_i$ and $y_i$ can be formulated as:

\begin{equation}
\begin{aligned}
\ell_{\text{align}}^{(i;time,topo)}= -\log \frac{\exp (s(z_i^a, y_i))}
    {\sum \limits_{j\in B} { ({\exp (s(z_i^a, y_j))}} + \mathbbm{1}_{[j\neq i]} {\exp (s(z_i^a, z_j^a))})}
\end{aligned}
\end{equation}
The final cross-modal contrastive loss is formulated as:

\begin{equation}
\mathcal{L}_{\text{cross}}=\frac{1}{2B} \sum_{i} \left(\ell_{\text {align}}^{(i;time,topo)}+\ell_{\text{align }}^{(i;topo,time)}\right)
\label{timetopoloss}
\end{equation}


\subsection{Overall Objective}
The proposed model jointly optimizes CL in time modality and time-topology correspondence. The overall loss function can be formulated as follows:
\begin{equation}
\mathcal{L}=\mathcal{L}_{time} + \alpha\mathcal{L}_{cross}
\label{eq:15}
\end{equation}
where $\alpha$ is a hyperparameter. 

Please note that instance discrimination using contrastive loss is not performed on topology modality. Instead, we apply contrastive loss to the time modality and leverage topology modality to enhance time series representation learning. The assumption is that time-topology alignment allows learning comprehensive and complementary semantic information by incorporating temporal features and their corresponding topological features.

\section{Experiments}
In this section, we conduct experiments on anomaly detection, classification, forecasting, and transfer learning tasks by applying TopoCL to state-of-the-art methods. Implementation details are presented in the appendix.

\subsection{Anomaly Detection}
Anomaly detection is used for identifying and addressing unusual patterns or behaviors across various domains. In this task, Yahoo~\cite{SR} and KPI~\cite{yahoo} datasets are used. We apply TopoCL to TS2Vec~\cite{yue2022ts2vec} and the evaluation is performed under normal and cold-start settings. For the normal setting,  each sample is divided into two portions: one for training and the other for testing. For the cold-start setting, the model is pretrained on the $FordA$ dataset from the UCR archive before evaluating on each target dataset. Following the evaluation protocol proposed by \cite{SR}, we determine whether the last point is an anomaly.

SPOT, DSPOT~\cite{POT}, DONUT~\cite{DONUT}, SR, and TS2vec methods are used as baselines for the normal setting, while FFT~\cite{FFT}, Twitter-AD~\cite{twitter}, and Luminol~\cite{luminol}, SR, and TS2vec are used as baselines for the cold-start setting. As shown in Table \ref{tab:anomaly}, in the normal setting, TopoCL improves the F$_1$ score by 1.9\% and 2.2\% on the Yahoo and KPI dataset, respectively. In the cold-start setting, our model outperforms the best baseline, achieving a 1.8\% higher F$_1$ score on the Yahoo dataset and a 0.5\% improvement on the KPI dataset. 

\begin{table}[h]
    \caption{Anomaly detection results of the proposed model and baselines on Yahoo and KPI datasets.}

  \centering
  \captionsetup{justification=centering}

  \scalebox{1}{
  \begin{tabular}{lcccccccc}
  \toprule 
    & \multicolumn{3}{c}{Yahoo} & \multicolumn{3}{c}{KPI} \\
    \cmidrule(r){2-4} \cmidrule(r){5-7}
    & F$_1$ & Prec. & Rec. & F$_1$ & Prec. & Rec. \\
    \midrule
    SPOT & 0.338 & 0.269 & 0.454 & 0.217 & 0.786 & 0.126\\
    DSPOT & 0.316 & 0.241 & 0.458 & 0.521 & 0.623 & 0.447 \\
    DONUT & 0.026 & 0.013 & 0.825 & 0.347 & 0.371 & 0.326 \\
    SR & 0.563 & 0.451 & 0.747 & 0.622 & 0.647 & 0.598\\
    TS2Vec & 0.745 & 0.729 & 0.762 & 0.677 & 0.929 & 0.533 \\
    TS2Vec+TopoCL & \textbf{0.764} & 0.763 & 0.764 & \textbf{0.699} & 0.924 & 0.562 \\
    \midrule
    \multicolumn{5}{l}{\textit{Cold-start:}}\\
    FFT & 0.291 & 0.202 & 0.517 & 0.538 & 0.478 & 0.615 \\
    Twitter-AD & 0.245 & 0.166 & 0.462 & 0.330 & 0.411 & 0.276 \\
    Luminol & 0.388 & 0.254 & 0.818 & 0.417 & 0.306 & 0.650 \\
    SR & 0.529 & 0.404 & 0.765 & 0.666 & 0.637 & 0.697 \\
    TS2Vec & 0.726 & 0.692 & 0.763 & 0.676 & 0.907 & 0.540 \\
    TS2Vec+TopoCL & \textbf{0.744} & 0.725 & 0.764 & \textbf{0.681} & 0.912 & 0.543 \\
    \bottomrule
  \end{tabular}
 }
  \label{tab:anomaly}
\end{table}

\subsection{Classification}
In this section, we conduct a classification task to evaluate the performance of the proposed model using the UCR \cite{UCRArchive2018} and UEA \cite{UEAArchive} archives, selecting 125 univariate and 29 multivariate time series datasets from each archive. We train an RBF kernel based SVM classifier on top of the learned representation, following the evaluation protocol used in T-Loss ~\cite{TLoss}. The penalty factor for the RBF kernel is  is selected by performing a grid search on the validation set, exploring the range $\{10^i \mid i \in [-4, 4]\}$

We apply TopoCL to TS2vec and compare its peformance with baselines including DTW, TST ~\cite{TST}, TS-TCC ~\cite{TSTCC}, T-Loss ~\cite{TLoss}, TNC ~\cite{TNC}, TS2vec, and InfoTS~\cite{luo2023time}. The classification results for the these datasets are summarized in Table \ref{tab:classification}. The classification accuracy of the proposed model outperforms the best baseline by 3.8\% on the UCR datasets and by 1.4\% on the UEA datasets. Detailed classification results are provided in the appendix.
\begin{table*}[h] 
    \centering
    \caption{Classification results of the proposed model and baselines on UCR and UEA datasets.}
    \resizebox{1\textwidth}{!}{
  \begin{tabular}{cccccccccccc}
  \toprule
    \multicolumn{2}{c}{Methods}  & TS2vec+TopoCL & TS2Vec & InfoTS & T-Loss & TNC & TS-TCC & TST & DTW \\
    \midrule
    \multicolumn{1}{c|}{\multirow{2}[2]{*}{29 UEA datasets}}  & \multicolumn{1}{c|}{Avg. Acc.} & \textbf{0.748 (+2.6\%)}  & 0.712  & 0.722  & 0.675 & 0.677 & 0.682  & 0.635 & 0.650 \\
    \multicolumn{1}{c|}{} & \multicolumn{1}{c|}{Avg. Rank} & \textbf{2.293}& 3.948 & 3.293 & 4.638 & 5.534 & 5.017 & 5.983 & 5.293\\
     \midrule
     \multicolumn{1}{c|}{\multirow{2}[2]{*}{125 UCR datasets}}  & \multicolumn{1}{c|}{ Avg. Acc.} & \textbf{0.853 (+1.5\%)} & 0.838  & 0.838  & 0.806 & 0.761  & 0.757 & 0.641 & 0.727\\
    \multicolumn{1}{c|}{} & \multicolumn{1}{c|}{Avg. Rank} & \textbf{2.004} & 3.048 & 2.392 & 4.540 & 5.376 & 5.268 & 7.156 & 6.216 \\
    \bottomrule
  \end{tabular}
}
    \label{tab:classification}
\end{table*}

\begin{figure}[h]
  \centering
    \includegraphics[width=0.7\linewidth]{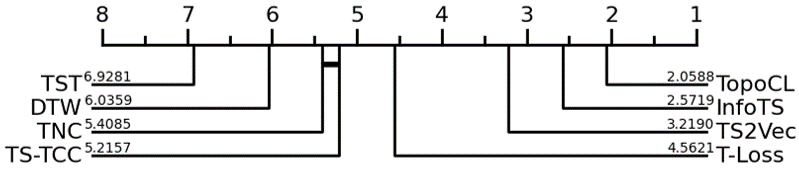}
  \caption{A Critical Difference (CD) diagram of representation learning methods on time series classification tasks, using 125 UCR datasets and 29 UEA datasets} \label{fig:rank}
\end{figure}

To further evaluate the effectiveness of TopoCL, we perform a significance test on the classification results. In the Figure \ref{fig:rank}, Critical Difference diagram ~\cite{CDDiagram} for Nemenyi tests on all datasets is presented. The Critical Difference diagram indicates that methods connected by a bold line do not have significantly different average ranks. From the results, we can conclude that the proposed model significantly outperforms other methods in average ranks. 

\subsection{Forecasting}
In this task, we use ETT~\cite{Informer} dataset and apply TopoCL to CoST~\cite{woo2022cost} and TS2Vec. We split the entire length of the time series into 60\% for training, 20\% for validation, and the remaining 20\% for testing. For evaluation, a ridge regressor with $L_2$ regularization is trained on top of the representation of the last time stamp $r_t$, to forecast the future timestamps. Specifically, we predict the future $H$ $\in$ {24, 48, 168, 720} observations based on the past $H$ time points. The validation set is used for selecting regularization penalty factor, exploring a search space of \{0.1, 0.2, 0.5, 1, 2, 5, 10, 20, 50, 100, 200, 500, 1000\}.

We use Informer~\cite{Informer},  LogTrans~\cite{LogTrans}, TCN~\cite{TCN}, LSTnet~\cite{LSTnet}, TS2Vec, and CoST as baselines, with mean absolute error (MAE) and mean squared error (MSE) serving as performance metrics. The evaluation results for univariate and multivariate forecasting are shown in Table \ref{tab:uniforecasting_full} and \ref{tab:multiforecasting_full}, respectively. In general, CoST + TopoCL outperforms the baselines in most cases, achieving reductions of 1.94\% in univariate forecasting and 1.76\% in multivariate forecasting, in terms of MSE. Furthermore, TS2Vec + TopoCL demonstrates a 3.39\% and 3.33\% decrease in average MSE compared to TS2Vec alone for univariate and multivariate forecasting, respectively.

\begin{table*}[t!]
    \caption{Univariate forecasting results.}
  \centering
      \resizebox{1\textwidth}{!}{
    \begin{tabular}{lrrrrrrrrrrrrrrrrr}
    \toprule
          &       & \multicolumn{2}{c}{CoST+} & \multicolumn{2}{c}{CoST} & \multicolumn{2}{c}{TS2Vec+} & \multicolumn{2}{c}{TS2vec} & \multicolumn{2}{c}{Informer} & \multicolumn{2}{c}{LogTrans} & \multicolumn{2}{c}{TCN} & \multicolumn{2}{c}{LSTnet} \\
    \midrule
    Metrics &       & \multicolumn{1}{l}{MSE} & \multicolumn{1}{l}{MAE} & \multicolumn{1}{l}{MSE} & \multicolumn{1}{l}{MAE} & \multicolumn{1}{l}{MSE} & \multicolumn{1}{l}{MAE} & \multicolumn{1}{l}{MSE} & \multicolumn{1}{l}{MAE} & \multicolumn{1}{l}{MSE} & \multicolumn{1}{l}{MAE} & \multicolumn{1}{l}{MSE} & \multicolumn{1}{l}{MAE} & \multicolumn{1}{l}{MSE} & \multicolumn{1}{l}{MAE
} & \multicolumn{1}{l}{MSE} & \multicolumn{1}{l}{MAE} \\
    \midrule
    \multicolumn{1}{c|}{\multirow{5}[2]{*}{\begin{sideways}ETTh1\end{sideways}}} & \multicolumn{1}{r|}{24} & \textbf{0.038} & \textbf{ 0.149} & 0.040 & 0.151 & 0.039 & 0.151 & 0.039 & 0.151 & 0.098 & 0.247 & 0.103 & 0.259 & 0.104 & 0.254 
& 0.108& 0.284\\
    \multicolumn{1}{c|}{} & \multicolumn{1}{r|}{48} & \textbf{0.057} & \textbf{0.182} & 0.060 & 0.186 & 0.062 & 0.189 & 0.062 & 0.190 & 0.158 & 0.319 & 0.167 & 0.328 & 0.206 & 0.366 
& 0.175& 0.424\\
    \multicolumn{1}{c|}{} & \multicolumn{1}{r|}{168} & \textbf{0.093} & \textbf{0.231} & 0.097 & 0.236 & 0.140 & 0.300 & 0.150 & 0.301 & 0.183 & 0.346 & 0.207 & 0.375 & 0.462 & 0.586 
& 0.396& 0.504\\
    \multicolumn{1}{c|}{} & \multicolumn{1}{r|}{336} & \textbf{0.107} & \textbf{0.252} & 0.112 & 0.258 & 0.172 & 0.329 & 0.174 & 0.332 & 0.222 & 0.387 & 0.230 & 0.398 & 0.422 & 0.564 
& 0.468& 0.593\\
    \multicolumn{1}{c|}{} & \multicolumn{1}{r|}{720} & \textbf{0.165} & \textbf{0.327} & 0.166 & 0.328 & 0.178 & 0.343 & 0.186 & 0.354 & 0.269 & 0.435 & 0.273 & 0.463 & 0.438 & 0.578 
& 0.659& 0.766\\
    \midrule
    \multicolumn{1}{c|}{\multirow{5}[2]{*}{\begin{sideways}ETTh2\end{sideways}}} & \multicolumn{1}{r|}{24} & \textbf{0.078} & \textbf{0.206} & 0.079 & 0.207 & 0.090 & 0.223 & 0.091 & 0.229 & 0.093 & 0.240 & 0.102 & 0.255 & 0.109 & 0.251 
& 3.554& 0.445\\
    \multicolumn{1}{c|}{} & \multicolumn{1}{r|}{48} & \textbf{0.116} & \textbf{0.257} & 0.118 & 0.259 & 0.123 & 0.271 & 0.124 & 0.272 & 0.155 & 0.314 & 0.169 & 0.348 & 0.147 & 0.302 
& 3.190& 0.474\\
    \multicolumn{1}{c|}{} & \multicolumn{1}{r|}{168} & 0.185 & \textbf{0.335} & 0.189 & 0.339 & \textbf{0.184} & 0.351 & 0.195 & 0.351 & 0.232 & 0.389 & 0.246 & 0.422 & 0.209 & 0.366 
& 2.800& 0.595\\
    \multicolumn{1}{c|}{} & \multicolumn{1}{r|}{336} & 0.204 & 0.359 & 0.206 & 0.360 & \textbf{0.202} & \textbf{0.362} & 0.203 & \textbf{0.362} & 0.263 & 0.417 & 0.267 & 0.437 & 0.237 & 0.391 
& 2.753& 0.738\\
    \multicolumn{1}{c|}{} & \multicolumn{1}{r|}{720} & 0.214 & 0.372 & 0.215 & 0.372 & \textbf{0.205} & \textbf{0.369} & 0.208 & 0.371 & 0.277 & 0.431 & 0.303 & 0.493 & 0.200 & 0.367 
& 2.878& 1.044\\
    \midrule
    \multicolumn{1}{c|}{\multirow{5}[2]{*}{\begin{sideways}ETTm1\end{sideways}}} & \multicolumn{1}{r|}{24} & \textbf{0.014} & \textbf{0.088} & \textbf{0.014} & \textbf{0.088} & \textbf{0.014} & 0.089 & 0.015 & 0.091 & 0.030 & 0.137 & 0.065 & 0.202 & 0.027 & 0.127 
& 0.090& 0.206\\
    \multicolumn{1}{c|}{} & \multicolumn{1}{r|}{48} & \textbf{0.025} & \textbf{0.117} & \textbf{0.025} & 0.118 & \textbf{0.025} & 0.118 & 0.028 & 0.126 & 0.069 & 0.203 & 0.078 & 0.220 & 0.040 & 0.154 
& 0.179& 0.306\\
    \multicolumn{1}{c|}{} & \multicolumn{1}{r|}{96} & \textbf{0.039} & \textbf{0.147} & \textbf{0.039} & 0.148 & 0.041 & 0.153 & 0.045 & 0.162 & 0.194 & 0.372 & 0.199 & 0.386 & 0.097 & 0.246 
& 0.272& 0.399\\
    \multicolumn{1}{c|}{} & \multicolumn{1}{r|}{288} & \textbf{0.074} & \textbf{0.205} & 0.078 & 0.212 & 0.093 & 0.231 & 0.096 & 0.237 & 0.401 & 0.554 & 0.411 & 0.572 & 0.305 & 0.455 
& 0.462& 0.558\\
    \multicolumn{1}{c|}{} & \multicolumn{1}{r|}{672} & \textbf{0.107} & \textbf{0.249} & 0.113 & 0.257 & 0.143 & 0.289 & 0.147 & 0.295 & 0.512 & 0.644 & 0.598 & 0.702 & 0.445 & 0.576 
& 0.639& 0.697\\
    \midrule
    Avg.  &       & \textbf{0.101} & \textbf{0.232} & 0.103 & 0.235 & 0.114 & 0.251 & 0.118 & 0.255 & 0.210 & 0.362 & 0.228 & 0.391 & 0.230 & 0.372 & 1.241& 0.535\\
    \bottomrule
    \end{tabular}
  \label{tab:uniforecasting_full}}
\end{table*}

\begin{table*}[]
  \caption{Multivariate forecasting results.}
  \centering
      \resizebox{\textwidth}{!}{

    \begin{tabular}{lrrrrrrrrrrrrrrrrr}
    \toprule
          &       & \multicolumn{2}{c}{CoST+} & \multicolumn{2}{c}{CoST} & \multicolumn{2}{c}{TS2Vec+} & \multicolumn{2}{c}{TS2vec} & \multicolumn{2}{c}{Informer} & \multicolumn{2}{c}{LogTrans} & \multicolumn{2}{c}{TCN} & \multicolumn{2}{c}{LSTnet} \\
    \midrule
    Metrics &       & \multicolumn{1}{l}{MSE} & \multicolumn{1}{l}{MAE} & \multicolumn{1}{l}{MSE} & \multicolumn{1}{l}{MAE} & \multicolumn{1}{l}{MSE} & \multicolumn{1}{l}{MAE} & \multicolumn{1}{l}{MSE} & \multicolumn{1}{l}{MAE} & \multicolumn{1}{l}{MSE} & \multicolumn{1}{l}{MAE} & \multicolumn{1}{l}{MSE} & \multicolumn{1}{l}{MAE} & \multicolumn{1}{l}{MSE} & \multicolumn{1}{l}{MAE} & \multicolumn{1}{l}{MSE} & \multicolumn{1}{l}{MAE} \\
    \midrule
    \multicolumn{1}{c|}{\multirow{5}[2]{*}{\begin{sideways}ETTh1\end{sideways}}} & \multicolumn{1}{r|}{24} & \textbf{0.383} & \textbf{0.428} & 0.384 & \textbf{0.428} & 0.560 & 0.521 & 0.588 & 0.531 & 0.577 & 0.549 & 0.686 & 0.604 & 0.583 & 0.547 & 1.293 & 0.901 \\
    \multicolumn{1}{c|}{} & \multicolumn{1}{r|}{48} & \textbf{0.432} & \textbf{0.459} & 0.437 & 0.464 & 0.597 & 0.544 & 0.624 & 0.556 & 0.685 & 0.625 & 0.766 & 0.757 & 0.670 & 0.606 & 1.456 & 0.960 \\
    \multicolumn{1}{c|}{} & \multicolumn{1}{r|}{168} & \textbf{0.637} & \textbf{0.580} & 0.643 & 0.582 & 0.734 & 0.631 & 0.764 & 0.639 & 0.931 & 0.752 & 1.002 & 0.846 & 0.811 & 0.680 & 1.997 & 1.214 \\
    \multicolumn{1}{c|}{} & \multicolumn{1}{r|}{336} & \textbf{0.793} & \textbf{0.671} & 0.812 & 0.679 & 0.896 & 0.717 & 0.937 & 0.731 & 1.215 & 0.896 & 1.397 & 1.291 & 1.132 & 0.815 & 2.143 & 1.380 \\
    \multicolumn{1}{c|}{} & \multicolumn{1}{r|}{720} & \textbf{0.916} & \textbf{0.754} & 0.970 & 0.771 & 1.054 & 0.799 & 1.066 & 0.800 & 1.215 & 0.896 & 1.397 & 1.291 & 1.165 & 0.813 & 2.143 & 1.380 \\
    \midrule
    \multicolumn{1}{c|}{\multirow{5}[2]{*}{\begin{sideways}ETTh2\end{sideways}}} & \multicolumn{1}{r|}{24} & 0.436 & 0.498 & 0.463 & 0.512 & \textbf{0.413} & \textbf{0.482} & 0.423 & 0.489 & 0.720 & 0.665 & 0.828 & 0.750 & 0.935 & 0.754 & 2.742 & 1.457 \\
    \multicolumn{1}{c|}{} & \multicolumn{1}{r|}{48} & 0.681 & 0.633 & 0.711 & 0.648 & \textbf{0.618} & 0.607 & 0.619 & \textbf{0.605} & 1.457 & 1.001 & 1.806 & 1.034 & 1.300 & 0.911 & 3.567 & 1.687 \\
    \multicolumn{1}{c|}{} & \multicolumn{1}{r|}{168} & \textbf{1.561} & \textbf{0.986} & 1.583 & 0.997 & 1.793 & 1.064 & 1.845 & 1.074 & 3.489 & 1.515 & 4.070 & 1.681 & 4.017 & 1.579 & 3.242 & 2.513 \\
    \multicolumn{1}{c|}{} & \multicolumn{1}{r|}{336} & \textbf{1.795} & \textbf{1.062} & 1.809 & 1.064 & 2.136 & 1.182 & 2.194 & 2.197 & 2.723 & 1.340 & 3.875 & 1.763 & 3.460 & 1.456 & 2.544 & 2.591 \\
    \multicolumn{1}{c|}{} & \multicolumn{1}{r|}{720} & \textbf{1.982} & \textbf{1.105} & 1.989 & 1.100 & 2.480 & 1.319 & 2.636 & 1.370 & 3.467 & 1.473 & 3.913 & 1.552 & 3.106 & 1.381 & 4.625 & 3.709 \\
    \midrule
    \multicolumn{1}{c|}{\multirow{5}[2]{*}{\begin{sideways}ETTm1\end{sideways}}} & \multicolumn{1}{r|}{24} & \textbf{0.244} & \textbf{0.327} & 0.246 & 0.330 & 0.438 & 0.435 & 0.448 & 0.440 & 0.323 & 0.369 & 0.419 & 0.412 & 0.363 & 0.397 & 1.968 & 1.170 \\
    \multicolumn{1}{c|}{} & \multicolumn{1}{r|}{48} & \textbf{0.329} & \textbf{0.384} & 0.334 & 0.389 & 0.574 & 0.513 & 0.595 & 0.524 & 0.494 & 0.503 & 0.507 & 0.583 & 0.542 & 0.508 & 1.999 & 1.215 \\
    \multicolumn{1}{c|}{} & \multicolumn{1}{r|}{96} & \textbf{0.376} & \textbf{0.417} & 0.382 & 0.423 & 0.613 & 0.543 & 0.629 & 0.551 & 0.678 & 0.614 & 0.768 & 0.792 & 0.666 & 0.578 & 2.762 & 1.542 \\
    \multicolumn{1}{c|}{} & \multicolumn{1}{r|}{288} & \textbf{0.470} & \textbf{0.484} & 0.477 & 0.489 & 0.673 & 0.585 & 0.690 & 0.594 & 1.056 & 0.786 & 1.462 & 1.320 & 0.991 & 0.735 & 1.257 & 2.076 \\
    \multicolumn{1}{c|}{} & \multicolumn{1}{r|}{672} & \textbf{0.621} & \textbf{0.574} & 0.625 & 0.576 & 0.766 & 0.644 & 0.780 & 0.652 & 1.192 & 0.926 & 1.669 & 1.461 & 1.032 & 0.756 & 1.917 & 2.941 \\
    \midrule
    Avg.  &       & \textbf{0.777} & \textbf{0.624} & 0.791 & 0.630 & 0.956 & 0.706 & 0.989 & 0.784 & 1.348 & 0.861 & 1.638 & 1.076 & 1.385 & 0.834 & 2.377 & 1.782 \\
    \bottomrule
    \end{tabular}
      \label{tab:multiforecasting_full}%
}
  \end{table*}%

\subsection{Transfer Learning}
Transfer learning on time series data brings a unique challenge, stemming from the complex nature of the data such as rapidly changing trends and shifts in temporal dynamics. In this section, we conduct transfer learning, following experiment settings proposed by TF-C \cite{zhang2022self}. 

The model is pretrained on SleepEEG dataset and finetuned on Epilepsy \cite{andrzejak2001indications-EPILEPSY}, Gesture \cite{liu2009uwave-Gesture}, FD-B \cite{lessmeier2016condition-FD-B}, and EMG datasets \cite{physiobank2000physionet-EMG}. We apply TopoCL to TS-TCC \cite{TSTCC} and compare our approach against several baselines: TS-TCC, TS2vec, CoST, Ti-MAE \cite{li2023ti-TiMAE}, TF-C, and TST. Performance is evaluated using accuracy and F$_1$ score. As shown in Table \ref{tab:transfer_topocl}, TS-TCC combined with TopoCL results in a notable improvement. %

\begin{table*}[htbp]
  \centering
\caption{Transfer learning results}
      \resizebox{0.65\textwidth}{!}{
    \begin{tabular}{lrrrrrrrr}
    \toprule
          & \multicolumn{2}{c}{Epilepsy} & \multicolumn{2}{c}{FD-B} & \multicolumn{2}{c}{Gesture} & \multicolumn{2}{c}{EMG} \\
    \midrule
          & \multicolumn{1}{l}{Acc.} & \multicolumn{1}{l}{F1} & \multicolumn{1}{l}{Acc.} & \multicolumn{1}{l}{F1} & \multicolumn{1}{l}{Acc.} & \multicolumn{1}{l}{F1} & \multicolumn{1}{l}{Acc.} & \multicolumn{1}{l}{F1} \\
    \midrule
    TS2Vec & 0.8395 & 0.9045 & 0.4790 & 0.4389 & 0.6917 & 0.6570 & 0.7854 & 0.6766 \\
    CoST  & 0.8840 & 0.7688 & 0.4706 & 0.3479 & 0.6833 & 0.6642 & 0.5365 & 0.3527 \\
    Ti-MAE & 0.7345 & 0.7720 & 0.6798 & 0.6336 & 0.7554 & 0.6932 & 0.6352 & 0.5832 \\
    TF-C  & 0.9495 & 0.9149 & 0.6938 & 0.7487 & 0.7642 & 0.7572 & 0.8171 & 0.7683 \\
    TST   & 0.8021 & 0.4011 & 0.4640 & 0.4134 & 0.6917 & 0.6601 & 0.4634 & 0.2111 \\
    TS-TCC & 0.9253 & 0.8633 & 0.5499 & 0.5418 & 0.7188 & 0.6984 & 0.7889 & 0.5904 \\
    TS-TCC+ & \textbf{0.9483} & \textbf{0.9133} & \textbf{0.7621} & \textbf{0.8174} & \textbf{0.7667} & \textbf{0.7605} & \textbf{1.0000} & \textbf{1.0000} \\
    \bottomrule
    \end{tabular}
\label{tab:transfer_topocl}}
\end{table*}%


\section{Analysis}
\subsection{Ablation Study}

In this section, we compare TopoCL to its five variants on 128 UCR datasets to study the effectiveness of the proposed components. The five variants are as follows: \textbf{(1) w/o time-topology alignment} removes the time-topology cross-modal alignment from TopoCL, \textbf{(2) w/o time domain contrastive loss} removes the instance and temporal contrast from $f^{topo}_\theta$, \textbf{(3) w/o $H_0$} excludes the 0-dimensional persistent homology when computing the persistence diagram, \textbf{(4) w/o $H_1$} excludes the 1-dimensional persistent homology, and \textbf{(5) avg-pool} replaces the symmetric aggregate function (max-pool) in $f^{topo}_\theta$ with avg-pool. As shown in Table ~\ref{tab:ablation}, any absence of the components lead to decrease in performance, showing that all components are imperative.

\begin{table}[h]
  \caption{Ablation results.}

  \centering
  \scalebox{1}{
  \begin{tabular}{p{5.5cm}c}
  \toprule
    & Avg. Accuracy \\
    \midrule
    \textbf{TopoCL} & \textbf{0.852} \\
    w/o time-topology alignment & 0.828 (-2.4\%) \\
    w/o time domain contrastive loss & 0.804 (-4.8\%) \\
    w/o $H_0$ & 0.829 (-2.3\%) \\
    w/o $H_1$ & 0.833 (-1.9\%) \\
    Avg-pool  & 0.845 (-0.7\%) \\
    \bottomrule
  \end{tabular}
  }
  \label{tab:ablation}
\end{table}
\subsection{Robustness under Data Augmentation}
\begin{figure}[h]
\includegraphics[width=0.5\textwidth]{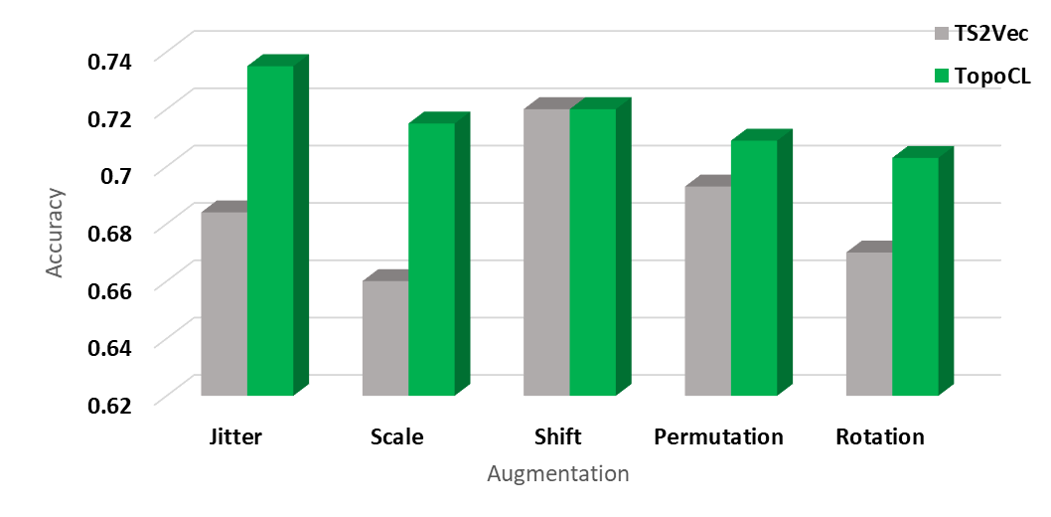}
\centering
\caption{Classification results on $Crop$ data with various augmentation techniques }
\label{fig:augmentation}
\end{figure}

We evaluate the robustness of TopoCL against various augmentation techniques using the $Crop$ dataset, which has the largest number of test samples in the UCR archive. Specifically, we individually apply data augmentation techniques, including jittering, scaling, shifting, permuting, and flipping.
As shown in Figure \ref{fig:augmentation}, TopoCL achieves steady performance compared to TS2vec. These results empirically validate that the time-topology alignment has a potential in compensating information loss during data augmentation process.

\subsection{Limited data scenario}
\begin{figure}[h]
\includegraphics[width=0.5\textwidth]{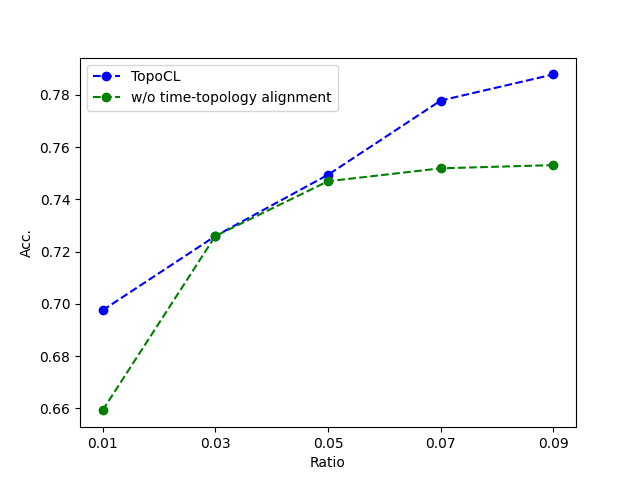}
\centering
\caption{Classification results in a limited data sample scenario. }
\label{fig:limited_data}
\end{figure}

Due to sensor malfunctions, data transmission errors, or manual data entry issues, sufficient amount of data are not always guaranteed. To verify the effectiveness of the proposed method in scenarios with limited data, we adjust the proportion of training samples. $FordB$ dataset from UCR archive is chosen for this analysis. The classification results using only 1\% to 9\% of the original training dataset are shown in Figure~\ref{fig:limited_data}. The proposed method consistently outperforms \textit{w/o time-topology alignment}. This result suggests that leveraging topological properties can capture the intrinsic characteristics of the data, even with a small amount of data.

\section{Conclusion}
This paper presents TopoCL, a topological contrastive learning for time series. We treat temporal and topological properties of time series as distinct modalities and propose a joint learning objective to enhance understanding of both, while also improving robustness to data augmentations. TopoCL is evaluated across multiple tasks: time series classification, forecasting, anomaly detection, and transfer learning. The experimental results demonstrate the universality, generalization capability, and effectiveness of the proposed model. Our ablations show that the joint learning of time modality contrastive loss and time-topology correspondences enhances these capabilities. Moreover, TopoCL shows consistent performance across various data augmentation techniques, suggesting its effectiveness in mitigating information loss resulting from these augmentations. In the future, we aim to extend TopoCL to accommodate large-scale and diverse pre-training datasets, thereby advancing toward a foundation model for time series analysis. 
However, calculating persistent homology from large-scale time series data requires significant computational costs. To address this, we aim to explore methodologies to reduce the increasing computational burden of persistent homology calculations.

\section*{Acknowledgments}
This work was supported by the National Research Foundation of Korea (NRF) Basic Research Lab Grant (No. 2021R1A4A1033486) and Midcareer Research Grant (No. 2020R1A2C2010200) by the South Korean government, as well as the KAIST Artificial Intelligence Graduate School Program (2019-0-00075).

\section*{Guidelines for Artificial Intelligence (AI)-Generated Text}
During the preparation of this work the authors used ChatGPT and Grammarly in order to refine the writing and check the grammar.




\bibliographystyle{IEEEtran}
\bibliography{IEEEabrv,Bibliography}

\ifCLASSOPTIONcaptionsoff
  \newpage
\fi



\end{document}